\documentclass[twocolumn]{template/fairmeta}

\usepackage{template/macro}
\usepackage{helvet}    
\usepackage{algorithm}
\usepackage{algorithmic}
\usepackage{newfloat}

\usepackage{listings}
\DeclareCaptionStyle{ruled}{labelfont=normalfont,labelsep=colon,strut=off} 
\definecolor{mygreen}{rgb}{0,0.6,0}
\definecolor{mygray}{rgb}{0.5,0.5,0.5}
\definecolor{lavender(floral)}{rgb}{0.71, 0.49, 0.86}
\definecolor{backcolor}{rgb}{1.0, 0.97, 0.91}

\lstdefinestyle{mystyle}{
  backgroundcolor=\color{backcolor!20},   
  basicstyle=\footnotesize\ttfamily,     
  breakatwhitespace=false,        
  breaklines=true,                 
  captionpos=b,                   
  commentstyle=\bfseries\color{green!50!black},    
  deletekeywords={...},            
  keywordstyle=\color{blue},       
  language=Python,                 
  numbers=none,
  columns=fullflexible,
  upquote=true,
  numbersep=5pt,                   
  numberstyle=\tiny\color{mygray}, 
  rulecolor=\color{black},         
  showspaces=false,                
  showstringspaces=false,         
  showtabs=false,                
  stringstyle=\color{lavender(floral)!75!black},     
  xleftmargin=4mm,
  tabsize=2,	                  
  title=\lstname                  
}

\floatstyle{ruled}
\newfloat{listing}{tb}{lst}{}
\floatname{listing}{Listing}

\usepackage{enumitem}

\usepackage{amsthm}

\definecolor{customgreen}{RGB}{61,157,115}
\definecolor{customred}{RGB}{213,112,95}
\definecolor{ocean}{rgb}{0.118, 0.392, 0.549}
\definecolor{bluebell}{rgb}{0.64, 0.64, 0.82}
\definecolor{aliceblue}{rgb}{0.94, 0.97, 1.0}
\definecolor{lavenderblue}{rgb}{0.8, 0.8, 1.0}

\def\usercol{customred}
\def\llmcol{ocean}
\def\datacol{bluebell}
\def\eodcol{customgreen}

\usepackage{mdframed}
\usepackage[inkscapelatex=false]{svg}

\newtheorem{thm}{Theorem}[section]
\newtheorem{lem}[thm]{Lemma}

\newtheorem{defn}[thm]{Definition}

\colorlet{shadebg}{black!5}

\mdfdefinestyle{thmstyle}{%
  topline   = false,
  bottomline= false,
  leftline  = false,
  rightline = false,
  backgroundcolor = shadebg,
  skipabove     = \baselineskip,
  skipbelow     = \baselineskip,
}

\surroundwithmdframed[style=thmstyle]{thm}
\surroundwithmdframed[style=thmstyle]{lem}
\surroundwithmdframed[style=thmstyle]{prop}
\surroundwithmdframed[style=thmstyle]{defn}
\surroundwithmdframed[style=thmstyle]{eg}
\surroundwithmdframed[style=thmstyle]{ex}
\surroundwithmdframed[style=thmstyle]{conj}
\surroundwithmdframed[style=thmstyle]{cor}
\surroundwithmdframed[style=thmstyle]{claim}
\surroundwithmdframed[style=thmstyle]{rmk}

\usepackage{booktabs}
\usepackage{multicol}
\usepackage{makecell}

\usepackage{tikz}

\usepackage{subcaption}

\usepackage{tcolorbox}

\title{Provable Benefits of In-Tool Learning for \\ Large Language Models}
\author[1,*]{Sam Houliston}
\author[2,*]{Ambroise Odonnat}
\author[3,*]{Charles Arnal}
\author[3,*]{Vivien Cabannes}

\affiliation[1]{ETH Zürich}
\affiliation[2]{Inria, Univ. Rennes 2 and IRISA}
\affiliation[3]{FAIR at Meta}

\contribution[\star]{Equal Contribution.}

\def\code{\url{https://github.com/ambroiseodt/itl}}

\abstract{Tool-augmented language models, equipped with retrieval, memory, or external APIs, are reshaping AI, yet their theoretical advantages remain underexplored.
In this paper, we address this question by demonstrating the benefits of \emph{in-tool learning} (external retrieval) over \emph{in-weight learning} (memorization) for factual recall. We show that the number of facts a model can memorize solely in its weights is fundamentally limited by its parameter count.
In contrast, we prove that tool-use enables unbounded factual recall via a simple and efficient circuit construction.
These results are validated in controlled experiments, where tool-using models consistently outperform memorizing ones.
We further show that for pretrained large language models, teaching tool-use and general rules is more effective than finetuning facts into memory.
Our work provides both a theoretical and empirical foundation, establishing why tool-augmented workflows are not just practical, but provably more scalable.}

\date{\today}
\correspondence{Vivien Cabannes at \email{vivc@meta.com}}

\begin{document}

\maketitle

\section{Introduction}
\label{sec:intro}

Large Language Models (LLMs) have revolutionized artificial intelligence, redefining how machines understand and generate human language.
Beyond this, LLMs are rapidly evolving from static predictors into dynamic, context-aware systems capable of reasoning, adapting, and acting over time.
Coding assistants are accelerating software development and lowering the barrier to entry for programming, hinting at the emergence of highly automated, agentic workflows.
This transformation is driven by advances in architecture and interaction design.
Retrieval-augmented generation \citep[RAG,][]{lewis2020rag} enables models to access external knowledge in real time, grounding their responses in contextually relevant information.
In parallel, memory augmentation, through the use of scratchpads or memory modules, empowers models to organize their reasoning, break down problems, iteratively refine outputs, and maintain coherence across extended sequences.

Together, these capabilities mark a shift away from purely in-weight learning, where all knowledge and reasoning must be encoded within the parameters of a model, toward more modular and interpretable systems that can utilize tools, access external information, and leverage structured memory.
Our study investigates the theoretical foundation for why such tool-augmented approaches outperform traditional monolithic models.
This evolution raises a fundamental question: 
What is the most efficient way for a model to acquire and utilize knowledge: should facts be internalized through parameter updates, or should models be taught to access and manipulate external sources of truth?
At the heart of this question lies a core tradeoff between in-weight learning, where information is compressed into the model's parameters during training, and tool-augmented learning (or in-tool learning), where the model learns to interact with external resources such as databases or APIs to retrieve information when needed.
While the former is bounded by the model’s capacity and sensitive to forgetting, the latter offers the potential for open-ended knowledge access, generalization, and interpretability.
In this work, we formalize this tradeoff and provide a rigorous theoretical framework for understanding the benefits of tool use in large language models.

\begin{figure*}[ht]
    \center
    \includegraphics[width=.9\linewidth]{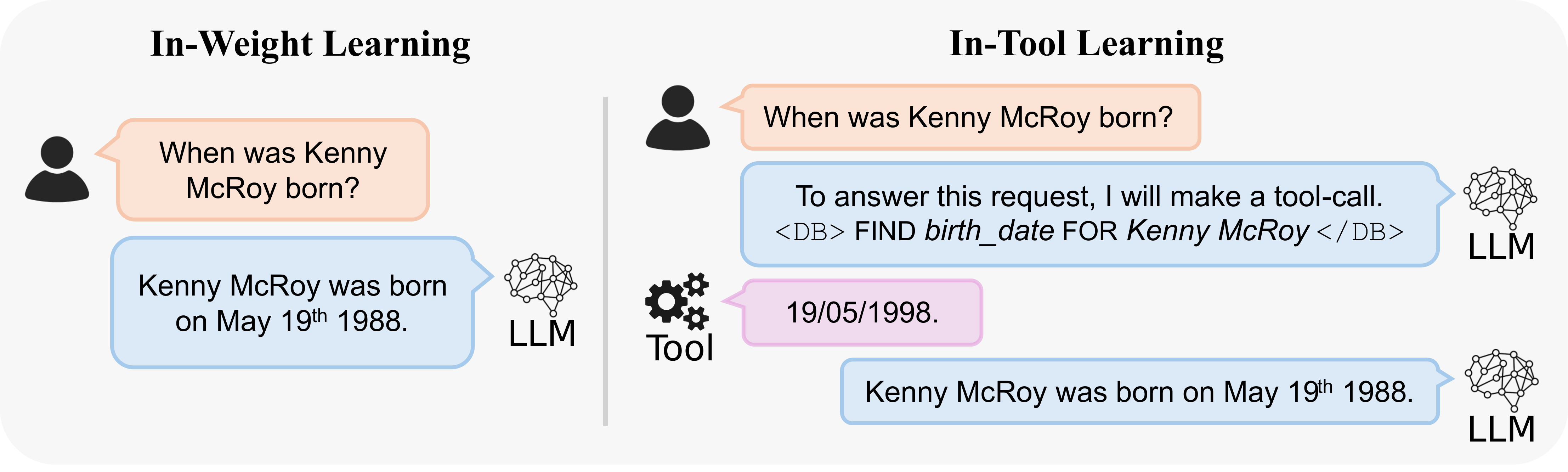}
    \caption{\textbf{Benefits of in-tool learning.} 
    Illustration of factual recall via \emph{in-weight learning} (memorization) versus \emph{in-tool learning} (external retrieval).
    While tool use may incur higher latency, it offloads storage, enabling scalable recall without increasing model size.
    It also better preserves the model’s original capabilities during finetuning.
    }
    \label{fig:intro}
\end{figure*}

\paragraph{\bf Contributions.} Our main contributions are as follows:

\begin{enumerate}[leftmargin=*]
    \item We derive a theoretical lower bound showing that the number of distinct facts a model can store purely in its weights is fundamentally limited by its number of parameters. This highlights a structural bottleneck in relying solely on in-weight memorization.
    \item We construct an explicit upper bound proving that tool-augmented models can, in principle, retrieve an unbounded number of facts by learning to interface with an external database. This result is supported by a formal circuit construction demonstrating the feasibility and efficiency of such tool use.
    \item We validate these theoretical results in a controlled experimental setting, where models are trained from scratch either to memorize facts or to learn to use an external tool. Our findings empirically confirm the predicted scaling laws and memory limitations.
    \item We demonstrate the practical implications of our findings for existing LLMs, showing that finetuning to learn new facts is significantly less effective than teaching models how to access structured tools or infer generalizable rules. This reinforces the argument that future LLM development should prioritize tool-use capabilities over rote memorization.
\end{enumerate}

Together, these results provide both a conceptual and empirical foundation for understanding why models that use tools outperform those that attempt to internalize all the knowledge.
Our study suggests a shift in model design philosophy: from training ever-larger monolithic models toward building modular systems that learn to query, not just to store information. Our codebase is freely available for researchers to further explore memory load in large language models\footnote{The code and reproducibility instructions are available at \code{}.}.

\paragraph{\bf Related work.}
Recent work has increasingly explored tool-augmented language models.
This trend is evident in the open-source literature, with models such as Toolformer~\citep{schick2023toolformerlanguagemodelsteach}, ReAct~\citep{yao2023reactsynergizingreasoningacting}, and HuggingGPT~\citep{shen2023hugginggptsolvingaitasks}, as well as in industrial systems that now support function calling, enabling modular LLM-agent architectures.
External memory mechanisms such as RAG~\citep{lewis2021retrievalaugmentedgenerationknowledgeintensivenlp}, RETRO~\citep{borgeaud2022improvinglanguagemodelsretrieving}, and retrieval-augmented transformers~\citep{shao2024scalingretrievalbasedlanguagemodels} serve as tools in their own right, enhancing knowledge access.
In a similar spirit, scratchpads~\citep{nye2021workscratchpadsintermediatecomputation} and chain-of-thought prompting~\citep{wei2023chainofthoughtpromptingelicitsreasoning} externalize reasoning and expand model capacity~\citep{besta2025reasoninglanguagemodelsblueprint}.
This is perhaps unsurprising, as several studies have quantified the limitations of in-weight knowledge~\citep{Hahn_2020,merrill2024expressivepowertransformerschain,cabannes2024iterationheadmechanisticstudy,zhu2025physics_scaling_law}, and highlighted interference during editing~\citep{meng2023locatingeditingfactualassociations,dai2022knowledgeneuronspretrainedtransformers}.

Taken together, these lines of research delimit a growing frontier that contrasts in-weight learning with increasingly sophisticated mechanisms for retrieval, reasoning, and tool use.
This is the dichotomy we aim to investigate in our work, contributing to the broader effort to deepen the theoretical and empirical understanding of language models~\citep[see, e.g.,][]{bietti2023birthtransformermemoryviewpoint,zekri2025largelanguagemodelsmarkov, beirami2025theoretical}. Additional related work can be found in Appendix~\ref{app:related_work} for the interested reader.

\section{Setting}
\label{sec:setting}
To highlight the differences between in-weight memorization and tool-augmented reasoning, we introduce a family of factual recall tasks.
Let ${\cal D}$ be a family of datasets, where each dataset $D \in {\cal D}$ is a finite collection of facts.
Each fact is a triplet $(n, a, v)$ of strings, consisting of:
\begin{itemize}
    \item a name $n \in {\cal N}$ (e.g., ``Thierry de Sienne''),
    \item an attribute $a \in {\cal A}$ (e.g., ``birthplace''), and
    \item a value $v \in {\cal V}_a$ (e.g., ``Germany'').
\end{itemize}
Each dataset $D$ defines a mapping $V_D: (n, a) \mapsto v$, which assigns to each name–attribute pair its corresponding value.
The goal of the recall task is to produce the ground-truth value $V_D(n, a)$ when given a query $(n, a)$.

Let ${\cal M}$ be a class of models, where each model $f \in {\cal M}$ implements a recall rule $R(f): {\cal N} \times {\cal A} \to \bigcup_{a \in {\cal A}} {\cal V}_a$.
For instance, the model class ${\cal  M}$ may be the set of all transformers with a given architecture, each model $f \in {\cal  M}$ corresponding to a specific choice of weights, and the recall rule $R(f)$ being defined by prompting the model with a query asking for the value associated to a pair $(n,a)$ and parsing the text obtained by auto-regressively sampling from the model until an end-of-sequence token is emitted.

\begin{defn}[Learnability]
    \label{def:learnability}
  We say that the model class ${\cal M}$ \emph{can learn} the recall task associated to a family of datasets ${\cal D}$ if, for each dataset $D \in {\cal D}$, there exists a model $f\in {\cal M}$ such that $f$ achieves perfect accuracy on the value recall task for $D$, i.e. $R(f) = V_D$. 
\end{defn}

Note that this definition abstracts away the process of training and does not account for sample efficiency or optimization difficulty.

\paragraph{\bf \bf Sentence format.}
Our study is concerned with factual recall performed by LLMs via question answering.
We define a dataset with a strict grammar that specifies the format of all question–answer pairs.

In the {\em in-weight} setting, the model is trained to directly generate the answer from its parameters.
The user produces a query string $Q$ using the format $Q = \phi_1(a) \circ \phi_2(n) \circ \phi_3(a)$, and the assistant must then generate the answer string $A$ in the form $A = \psi_1(n) \circ \psi_2(a) \circ \psi_3(v)$, where $\phi_i$ and $\psi_i$ are deterministic string functions (i.e. templates) that encode the structure of the input and output, and $\circ$ denotes string concatenation.
For example, given the previous fact example, we have:
\[
Q = \underbrace{\text{Where was}}_{\phi_1(a)}\ \underbrace{\text{Thierry de Sienne}}_{\phi_2(n) = n}\ \underbrace{\text{born?}}_{\phi_3(a)}
\]
\[
A = \underbrace{\text{Thierry de Sienne}}_{\psi_1(n) = n}\ \underbrace{\text{was born in}}_{\psi_2(a)}\ \underbrace{\text{Germany}}_{\psi_3(v) = v}
\]

In the {\em in-tool} setting, the model learns to issue a structured tool query that retrieves the value from an external database.
The user query $Q$ remains unchanged, but the assistant must first produce an intermediate tool query string $T = \chi_1(a) \circ \chi_2(n)$, where the $\chi_i$ are deterministic string templates encoding the attribute and name in a format expected by the tool interface.
This query is sent to an external retrieval system (e.g., a database or API), which returns the corresponding value $\xi(f, a, n)$, where $\xi(f, a, n)=v$ if the assistant correctly formats the query, and an error message otherwise.
The assistant then produces the final answer $A$, reusing the same output grammar as in the in-weight case.

For example, using the same fact example, we have: 
\begin{align*}
 T & = \underbrace{
        \text{\tiny To answer this request, I will make a tool-call. \footnotesize \texttt{\footnotesize <DB>} FIND birthplace}}_{\chi_1(a)} \\
& \qquad \underbrace{\text{\footnotesize FOR Thierry de Sienne \texttt{\footnotesize </DB>}}}_{\chi_2(n)}, 
\end{align*}
This formal grammar establishes a controlled setting to characterize the information storage and retrieval capabilities of LLMs precisely. An example of queries formatted in JSON files can be found in Appendix~\ref{app:data}.

\section{In-Weight Lower Bound}
\label{sec:lower_bound}

This section illustrates the limitations of in-weight learning through theoretical lower bounds.
The following lemma provides a lower bound on the size of a model class that can learn a given family of datasets.

\begin{lem}[Capacity lower bound.]\label{lem:lower-bound}
Let ${\cal D}$ be a family of factual datasets defined over a fixed set of names ${\cal N}$ and attributes ${\cal A}$.
Let ${\cal M}$ be a model class that can learn ${\cal D}$. Then, we have
\begin{equation}
    |{\cal M}| \geq |{\cal D} |.
\end{equation}
\end{lem}
\begin{proof}
    This follows from a simple enumeration argument, since the definition of learnability (Definition~\ref{def:learnability}) is equivalent to the inclusion $\{V_D \,\vert\, D\in{\cal D}\} \subseteq \{R(f) \,\vert\, f \in {\cal M}\}$. Considering the cardinality of those sets concludes the proof.
\end{proof}

Lemma~\ref{lem:lower-bound} implies a lower bound on the number of parameters required to define a class of models that could memorize a given set of facts. This allows us to provide an upper bound on the parameter requirement.

\begin{thm}[Parameter lower bound]\label{thm:lower-bound}
Let ${\cal M}$ be a class of models with $P$ parameters, each quantized to $b$ bits.
Let ${\cal D}$ be the family of all datasets defined over names ${\cal N}$, attributes ${\cal A}$, and value sets ${\cal V}_a$.
Then, if ${\cal M}$ can learn ${\cal D}$ without accessing external tools (in-weight learning), the number of parameters  must satisfy:
\[
    P \geq \frac{|{\cal N}|}{b} \sum_{a\in{\cal A}}\log_2 |{\cal V}_a| = c\cdot \# \text{Facts},
\]
where $\#\text{Facts} = |{\cal N} \times {\cal A}|$ is the number of facts, and $c$ is the average of $\log_2 |{\cal V}_a| / b$ over $a \in {\cal A}$.
\end{thm}
\begin{proof}
    This is a consequence of Lemma \ref{lem:lower-bound}.
    The total number of distinct parameter configurations in the model family ${\cal M}$ is at most $2^{bP}$ (i.e., $|\mathcal{M}| \leq 2^{bP}$). Moreover, defining any dataset from ${\cal D}$ requires distinguishing between all possible assignments of values for the attributes of each name. For a single name, there are $\prod_{a\in{\cal A}} |\mathcal{V}_a|$ possible value assignments; for all $\mathcal{N}$ names, this yields $|\mathcal{D}|=\left(\prod_{a\in{\cal A}} |\mathcal{V}_a|\right)^ \mathcal{N}$. Applying Lemma \ref{lem:lower-bound} with these quantities and taking the base-2 logarithm of both sides of the inequality completes the proof. 
\end{proof}

Theorem \ref{thm:lower-bound} shows that the number of parameters needed to memorize arbitrary fact mappings grows linearly with the number of facts.
It proves a hard capacity ceiling on in-weight learning: Once the number of facts exceeds a threshold determined by model size, memorization becomes impossible without architectural changes or external memory.

Note that Theorem \ref{thm:lower-bound} could be refined in two directions.
First, the capacity of a model class does not necessarily grow exponentially with the number of quantization bits. In practice, empirical studies such as \citet{zhu2025physics_scaling_law} suggest that the effective representational capacity of language models is often much lower than the theoretical maximum--typically around 2 bits per parameter, meaning that $b$ could be replaced by $\approx 2$ in the theorem.

Second, in practice, facts are often highly structured and correlated.
For example, knowing that someone is Japanese makes it more likely that they are older than 60 than if they were from Kenya, due to population-level age differences.
These kinds of regularities mean that not all combinations of facts are equally likely. 
One can account for this by introducing a probability distribution over datasets in ${\cal D}$, and then asking: How many parameters are needed for a model to correctly learn the most likely $(1 - \epsilon)$ fraction of datasets? 
However, even with these statement refinements, the relationship between the number of parameters and the number of learnable ``effective facts'' still scales linearly.


\section{In-Tool Upper Bound}
\label{sec:upper_bound}

In this section, we show that tool use allows models, and in particular transformers~\citep{vaswani2017attention}, to circumvent the limitations of in-weight learning and accurately recall an unlimited number of facts without needing additional parameters.
More specifically, we exhibit an upper bound on the number of parameters needed for a transformer to implement the kind of database querying described in Section~\ref{sec:setting}.

Let ${\cal D}$ be a family of factual datasets defined over some names, attributes, and value sets ${\cal N}, {\cal A}, \{{\cal V}_a\}_{a\in {\cal A}}$. 
We introduce the following notion of {\em query-based learnability} for a model augmented with access to an external database.

\begin{defn}[Query-based Learnability]
    \label{def:augmented}
    For a factual dataset $D$, the associated {\em retrieval system} $R_D$ is a recall rule which can be queried in the manner described in Section~\ref{sec:setting} to retrieve any fact of $D$.
    We say that a tool-augmented model $f$ solves the recall task for a family of datasets ${\cal D}$ if, for any $D \in {\cal D}$, it achieves perfect accuracy on the value recall task for $D$ by querying $R_D$.
\end{defn}

Note that Definition~\ref{def:augmented} is stronger than Definition~\ref{def:learnability}, as a single model (rather than a family of models) is tasked with ``learning'' any dataset $D\in \cal D$, provided it has access to a correct retrieval system.

In the theorem below, we consider Llama3-like transformers \citep{grattafiori2024llama3herdmodels} with a fixed token vocabulary; we make a few mild technical assumptions, e.g. the queries and answers are shorter than the maximum model context length, which we detail in Appendix~\ref{app:theory}, along with the proof of the statement.

\begin{thm}[Parameter upper bound]\label{thm:upper-bound}
Let ${\cal D}$ be a family of factual datasets defined over some names, attributes, and value sets ${\cal N}, {\cal A}, \{{\cal V}_a\}_{a\in {\cal A}}$. 
Given $\cal R_{\cal D} = \{R_D\}_{D \in {\cal D}}$ a family of retrieval systems associated to~$\cal D$, there exists a transformer with at most $8$ transformer blocks, an embedding dimension of at most $O(|{\cal A}|)$ and a total number of parameters at most $O(|{\cal A}|^2)$ that can solve the value recall task for $\cal D$ if equipped with the ability to access ${\cal R}_{\cal D}$.
\end{thm}

\begin{proof}[Proof sketch]
The proof consists of designing algorithmic mechanisms to auto-regressively generate the tool query $T$ and the answer $A$ in the string $Q \circ T \circ \xi \circ A$.
At a high level, the transformer needs to: ({\em i}) parse the attribute $a$ from the user query $Q$; ({\em ii}) emit the corresponding $\chi_1(a)$; ({\em iii}) emit the suffix $\chi_2(n)$ of $T$ by copying the name part from the question $Q$; and ({\em iv}) perform a similar copying operation to emit the final answer $A$ after observing the value $v$ returned by the database.

\emph{Parsing the attribute from $Q$.}
First, one could reserve the first transformer block to retrieve absolute position encoding \cite[see, e.g.][Proposition 2.2]{golkar2024contextualcountingmechanisticstudy}.
Then, one can reserve $2|\mathcal{A}|$ attention heads in the second block, so that each head fires whenever the suffix of the input stream matches one of the fixed strings $\phi_1(a)$ or $\phi_3(a)$.
The firing pattern can be written into a one-hot ``attribute register'' of dimension $2|\mathcal{A}|$ that is carried forward in the residual stream.

\emph{Emitting $\chi_1(a)$.}
The attribute $a$ can be retrieved from the attribute register, which in turn enables the transformer to auto-regressively output $\chi_1(a)$.

\emph{Emitting $\chi_2(n)$.}
The emission of $\chi_2(n)$ can be done by copying $\phi_2(n)$. This requires parsing $\phi_2(n)$, which can be done using the attribute register to identify the position of the token that starts $\phi_2(n)$, followed by using an induction head to copy the full name. A mechanism such as counting the number of whitespace characters in $Q$ and subtracting the number of whitespace characters in $\phi_1(a) \circ \phi_3(a)$ can be used to determine when to stop copying (thus avoiding the inclusion of $\phi_3(a)$ when emitting $\chi_2(n)$).

\emph{Emitting $A$.}
A similar construction allows copying the value $v$ returned by the database, as well as the name $n$, to emit the final answer $A$.

Overall, this construction uses embeddings of dimension $O(|\mathcal{A}|)$ with a constant number of layers, leading to a total number of parameters that scales as $O(|\mathcal{A}|^2)$.
\end{proof}

Note that the bounds do not depend on the number of names or possible values.
While our theoretical results provide foundational insights, empirical validation remains crucial, since Theorem~\ref{thm:lower-bound} describes a worst-case scenario, and Theorem~\ref{thm:upper-bound} is an existence result that does not account for optimization difficulty.


\section{Controlled Experiments}
\label{sec:controlled_exp}

This first experimental section investigates the difference between in-weight and in-tool learning by pretraining models from scratch on synthetic factual datasets. 
Our results support the theoretical insights: tool-augmented recall outperforms in-weight memorization in terms of the number of facts learned per parameter. Additional experimental details and results are provided in Appendix~\ref{app:controlled_exp}.

\begin{figure}[t]
    \centering
    \includegraphics{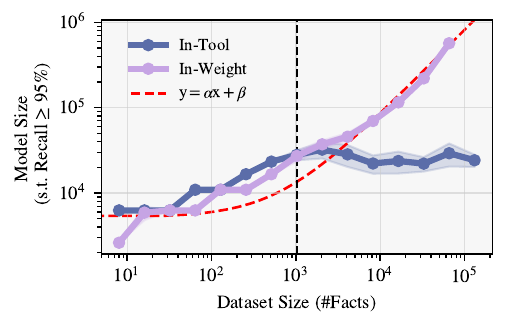}
    \caption{\textbf{Scaling of parameter requirements with the number of facts.} 
    Minimum number of parameters required to achieve at least a 95\% recall as a function of the dataset size. Results are averaged across 10 runs with standard deviation.
    In-weight models require increasingly more parameters as the number of facts grows, approximately following the linear trend $y = \alpha x + \beta$ with $(\alpha, \beta) = (8.14, 5171)$, consistent with Theorem~\ref{thm:lower-bound}.
    In contrast, in-tool models exhibit a sharp saturation: beyond a critical point (dashed vertical line around 1K facts), the parameter requirement flattens, indicating a transition to tool-based retrieval.}
    \label{fig:controlled_scaling}
\end{figure}

\begin{figure}[t]
    \centering
    \includegraphics{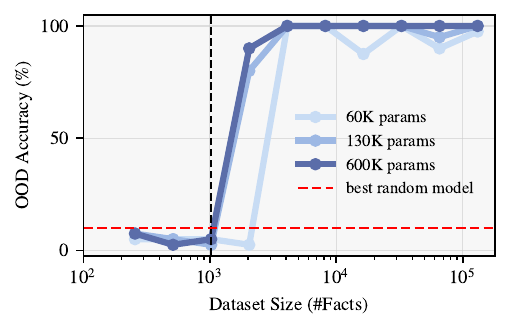}
    \caption{\textbf{Transition from memorization to retrieval in in-tool models.} 
    Accuracy on out-of-distribution databases (i.e., databases of facts unseen during training) as a function of the dataset size.
    Below the transition point (the same dashed vertical line as in Figure~\ref{fig:controlled_scaling}), in-tool models memorize specific databases and generalize poorly, worse than the best random model that outputs the same value for all queries (red dashed line). Above this threshold, they learn to construct tool queries that generalize across datasets, resulting in stable accuracy even on new data.} 
    \label{fig:controlled_phase}
\end{figure}

\paragraph{\bf Experimental setup.} To validate our theoretical findings, we place ourselves in a controlled setting. 
We construct synthetic biographical datasets from a fixed list of names ${\cal N}$ and four attributes ${\cal A}$: ``birth place'', ``birth date'', ``current address'', ``occupation''; which can take 7, 16800, 213, and 100 values respectively.
This yields $4|{\cal N}|$ atomic facts per dataset instance. For each dataset size, we train a family of small Llama3-style transformer models~\citep{grattafiori2024llama3herdmodels} with 2 layers, 2 attention heads, a vocabulary size of 260 (byte encoding with 4 special tokens), and a context window of 257. 
The embedding dimension ranges from 4 to 128, resulting in models with between 2K and 0.6M parameters. 
Our models are trained for up to $100{,}000$ steps using the AdamW optimizer~\citep{loshchilov2018decoupled}, with a batch size of $128$ samples, a decoupled weight decay coefficient of $0.1$, and $(\beta_1, \beta_2) = (0.9, 0.95)$.
We use a cosine learning rate scheduler with a warmup phase of $50$ steps and a maximum learning rate of $0.001$, and a final learning rate ratio of $0$.

We compare two training regimes, illustrated in Figure~\ref{fig:intro}: 
\begin{enumerate}[label=(\roman*)]
    \item {\em in-weight}, where the model must directly generate the answer from its parameters, and 
    \item {\em in-tool}, where the model first emits a structured lookup query to an external database, which returns the requested value that the model then formats into an answer.
\end{enumerate}

\paragraph{\bf Theory validation.}
Our empirical results strongly support the theoretical predictions presented in Sections~\ref{sec:lower_bound} and~\ref{sec:upper_bound}.
As shown in Figure~\ref{fig:controlled_scaling}, the in-weight regime exhibits an unbounded increase in the number of required parameters as the number of facts grows.
This behavior demonstrates the existence of a lower bound on parameter requirements, consistent with the general claim of Theorem~\ref{thm:lower-bound}.
Note that the empirical scaling is slightly sublinear for small dataset sizes, approximately following a square-root trend.
This suggests the presence of subtle phenomena not captured by the theorem, which we leave to future work.
This confirms that parameter count imposes a hard limit on memorization-based learning.
In contrast, the in-tool regime displays a clear transition: after a critical dataset size (around 1,000 facts in our setting), the parameter requirement saturates and remains nearly constant.
This plateau confirms that the model has shifted from memorizing facts to delegating recall to an external database, in line with the existence results in Theorem~\ref{thm:upper-bound}.

\paragraph{\bf From memorization to rule learning.}
This transition in the in-tool regime reflects more than a capacity shift; it reveals a learning phase transition.
Initially, models faced with tool-augmented recall resort to memorizing fact–answer pairs, similar to the in-weight regime. 
However, once exposed to a sufficient diversity of facts, the model discovers the underlying logic for constructing tool queries. As shown in Figure~\ref{fig:controlled_phase}, where the best OOD accuracy achieved by in-tool models is displayed, this results in a dramatic improvement in generalization to out-of-distribution facts.
This phenomenon mirrors the ``grokking'' effect observed in other settings: a delayed but sharp shift from brute-force memorization to systematic rule learning~\citep{power2022grokkinggeneralizationoverfittingsmall,nanda2023progress}.
Once the model learns the correct format for querying, retrieval generalizes across databases, decoupling recall accuracy from the number of stored facts.

\begin{figure}[t]
    \centering
    \includegraphics{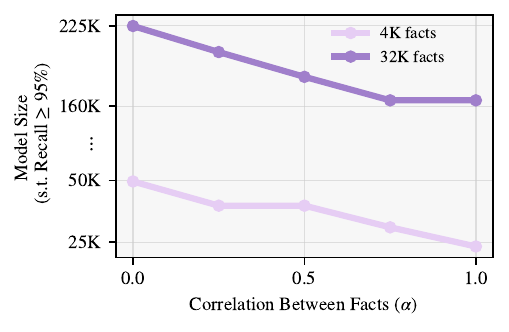}
    \caption{\textbf{Impact of the dependency between facts on in-weight memorization.}  Minimum parameter requirement to achieve at least a 95\% recall on datasets generated with a correlation between facts of $\alpha$. 
    The introduction of such a correlation breaks the independence between triplets, making the dataset more compressible thanks to common patterns between facts. Thus, as correlation increases, the number of “effective” units of facts decreases, requiring fewer parameters to store them.
    }
    \label{fig:controlled_compression}
\end{figure}

\paragraph{\bf What counts as a fact?}
Our study so far assumes a clean distinction between facts (e.g., ``Thierry de Sienne was born in Germany'') and rules (e.g., ``birthplace queries can be answered by a lookup function'').
In our attribution task, this boundary is sharp, but in broader domains such as mathematics or commonsense reasoning, the line between facts and rules is often ambiguous.
Are theorems facts, or derived consequences of axioms? 
Ontologies may reveal intricate interdependencies that blur this distinction.

To explore how structure affects memorization, we introduce controlled correlations between facts using a parameter $\alpha \in [0,1]$.
For each pair name-attribute $(n, a)$, we draw a Bernoulli variable $X \sim \text{Bern}(\alpha)$, and set $V_D(n, a) = V_1(n, a)$ if $X = 1$, or $V_2(n, a)$ otherwise, where, $V_1$ assigns values based on the family name, while $V_2$ assigns values independently at random.
As such, $\alpha = 1$ amounts to family members sharing a birthplace, current address, and occupation, while $\alpha = 0$ corresponds to the unstructured case used in earlier experiments.
As shown in Figure~\ref{fig:controlled_compression}, increasing the correlation reduces the number of parameters required for in-weight models to achieve high recall. 
This effect could be understood in light of Lemma~\ref{lem:lower-bound}: by introducing correlations, we constrain the set of possible datasets\footnote{This constraint applies in expectation: with high probability, the dataset is sampled from a smaller effective family, tightening the lower bound in expectation.}, thereby lowering the lower bound on model capacity.
Our results suggest that transformers are able to adapt to this structure, effectively compressing related facts through shared patterns and achieving high recall with fewer parameters.
This aligns with previous findings on how structural assumptions about the data influence learning in transformers and neural networks more broadly \citep{valvoda-etal-2022-benchmarking, DziriFaith2023, Arnal2024LearningWH,  WangGrokking2025,  mahajan2025compositional}.


\section{Large-Scale Experiments}
\label{sec:large_scale_exp}
This section investigates whether the limitations of in-weight learning persist in real-world settings involving pretrained multi-purpose language models. 
In particular, we show that while large models can store vast amounts of information in their parameters, introducing new knowledge through finetuning may interfere with existing capabilities due to finite capacity and training dynamics.
In contrast, tool-augmented learning promises scalability without forgetting. We provide further related results in Appendix~\ref{app:large_scale_exp}. 

\paragraph{\bf Experimental setup.}
We fine-tune instruction-tuned language models on the synthetic factual datasets from Section~\ref{sec:controlled_exp}, ranging from 500 to 50k atomic facts. 
Models include SmolLM 2 Instruct (135M, 360M, 1.7B)~\citep{allal2025smollm2smolgoesbig} and Llama 3.1/3.2 Instruct (1B, 3B, 8B)~\citep{grattafiori2024llama3herdmodels}, each trained to reach at least 95\% recall.
We use the same training setup as in Section~\ref{sec:controlled_exp}, with learning rates adapted to model size.
For SmolLM models, we use the values reported in the original paper: $10^{-3}$ for 135M and 360M, and $3 \cdot 10^{-4}$ for 1.7B. 
For Llama models, we scale learning rates $\eta$ with model size P, setting $\eta = 2 \cdot 10^{-5}$ at 8B, and following the established inverse power law  $\eta\propto P^{-1/2}$ for stable updates, with $P$ the number of parameters~\citep{kaplan2020scalinglawsneurallanguage,gao2022scalinglawsrewardmodel}.

Frequent checkpoints are saved to evaluate: 
(i) factual recall, 
(ii) HellaSwag accuracy, used as a proxy for general language abilities; this benchmark presents four candidate sentence completions, and the model is evaluated on its ability to select the most plausible one~\citep{zellers2019hellaswag}, and 
(iii) Total Variation (TV) distance from the base model’s output distribution, estimated over 100 natural language prompts by generating completions with the base model and computing the mean token-level $\ell_1$ distance between base and fine-tuned output probabilities.

\begin{figure}[t]
    \centering   \includegraphics{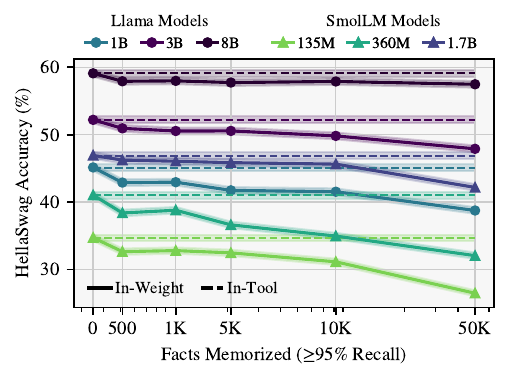}
    \caption{
    \textbf{Impact of factual memorization on general language capabilities (HellaSwag)}. Hellaswag accuracy as a function of the dataset size for pretrained models finetuned until reaching at least a 95\% recall on the database. Tool-learning (dashed) preserves general capabilities almost perfectly across models and dataset sizes. In contrast, in-weight learning (solid) leads to noticeable degradation, especially for smaller models and larger factual loads. This supports our theoretical prediction that parameter-based memorization causes interference and loss of prior capabilities, due to finite capacity and update interference. Larger models are more robust to such forgetting, but still exhibit performance decay beyond ~10k facts.}
    \label{fig:largescale_capability}
\end{figure}

\paragraph{\bf Overloading threatens general capabilities.}
Motivated by our theoretical results, we now investigate whether the benefits of in-tool learning persist in large, pretrained language models.
Figure~\ref{fig:largescale_capability} shows how HellaSwag accuracy evolves as models are trained to memorize increasing numbers of facts.
In-tool learning (dashed lines) retains general performance across all scales, confirming that externalizing factual storage prevents interference.
In contrast, in-weight learning (solid lines) shows a clear degradation, especially in smaller architectures.
This supports our theoretical intuition: parameter-based memorization consumes capacity and risks overwriting prior knowledge. 
We note that larger models degrade more slowly, suggesting they can absorb new information with less interference.
These results underscore the scalability advantage of tool-augmented learning for preserving general capabilities.

\begin{figure}[t]
    \centering
\includegraphics{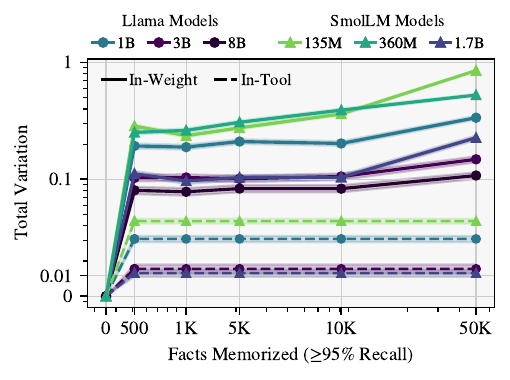}
    \caption{\textbf{TV distance for different memorization loads}. Similar setup to Figure~\ref{fig:largescale_capability}, where we display the Total Variation (TV) distance.
    In-weight finetuned models deviate more from their base models when memorizing bigger datasets.}
    \label{fig:largescale_distance}
\end{figure}
\paragraph{\bf Token distribution change.}
The differences between in-weight and in-tool learning can be further understood by measuring how much each method alters the model’s output distribution.
As shown in Figure~\ref{fig:largescale_distance}, in-weight learning leads to significant drift—especially in smaller models—with total variation (TV) distance growing sharply with the number of memorized facts.
In contrast, tool-learning causes minimal divergence, even at large scales.
This offers a complementary perspective on our results: in-weight learning requires shifting the next-token distribution to encode new information, while in-tool learning preserves the model’s original behavior by delegating recall.

\begin{figure}[t]
    \centering   
    \includegraphics{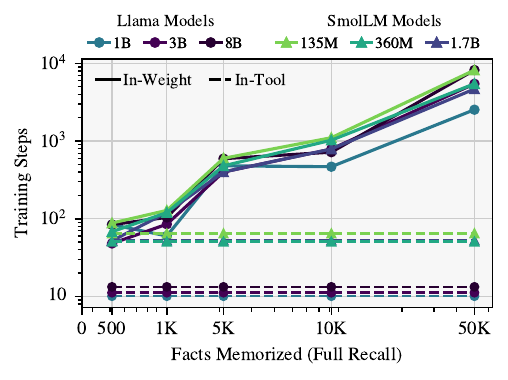}
    \caption{
    \textbf{Training steps required for memorization}. Number of training steps as a function of the dataset size for pretrained models finetuned until perfect recall. For the same optimization parameters, in-weight compute requirements scale with the memorization load. On the other hand, learning tool-use is faster and independent of dataset size in our setup. Llama models are better predisposed to learn tool-use from their pretraining compared to SmolLM models. 
    }
    \label{fig:largescale_efficiency}
\end{figure}
\paragraph{\bf Efficiency trade-offs.}
Another advantage of in-tool learning is its training efficiency: 
As shown in Figure~\ref{fig:largescale_efficiency}, models acquire the tool-use function call pattern in fewer than 20 steps, as soon as the set is large enough to elicit tool-use. In contrast, in-weight learning requires substantially more training to memorize individual facts.
However, this efficiency comes with a trade-off:
In-weight models support cheaper inference, producing answers directly without generating intermediate tool queries or relying on external calls.
We also observe that tool-use can be slightly unstable to train; early stopping proved effective to prevent capability loss once the tool behavior was acquired.


\section{Discussion}
\label{sec:discussion}

This work provides a unified theoretical and empirical account of the tradeoff between in-weight and in-tool learning.
Theoretically, we establish that the number of facts a model can store in its parameters is bounded by its size, implying that scaling knowledge capacity through model enlargement is inherently inefficient.
In contrast, models that learn to interact with external tools can access unbounded factual knowledge without increasing their parameter count.
Controlled experiments confirm this sharp divide: while in-weight models require ever-larger architectures to memorize growing datasets, tool-augmented models exhibit a phase transition, rapidly shifting to rule-based querying once sufficient diversity is observed.
This decouples memory capacity from model size.
Large-scale experiments extend these findings to pretrained models, showing that in-weight finetuning for factual recall degrades general capabilities, a consequence of limited capacity forcing new information to overwrite prior knowledge.
Tool-based approaches, by externalizing factual storage, preserve core skills, reduce training costs, and introduce minimal behavioral drift.
These results underscore a key design insight: language models scale more effectively not by internalizing ever more information, but by learning to access and orchestrate it, emerging less as static predictors and more as systems capable of modular interaction with structured external resources.

\paragraph{\bf Limitations.}
While our analysis provides conceptual clarity and theoretical insight, it is grounded in idealized, controlled settings. Although insightful, such synthetic datasets may not fully reflect the complexity of real-world knowledge. 
Moreover, the optimization dynamics are not considered in our theoretical bounds, and the exploration of tool use is limited to factual recall via structured databases, excluding other important tools such as computer use, reasoning trace generation, or learnable memory modules integrated into the model architecture.
Extending our analysis to these richer settings is a promising direction for future work. 


\bibliographystyle{template/arxiv}
\bibliography{references}


\clearpage
\appendix
\onecolumn
\textbf{\LARGE Appendix}
\section{Extended Related Work}
\label{app:related_work}
\paragraph{\bf Memorization capabilities of language models.}
The "Physics of LLMs" series (Parts 3.1–3.3) offers a comprehensive and theoretically grounded account of various aspects of memorization with LLMs. Part 3.1~\citep{PhysicsLLMsPart3.1} covers knowledge storage, distinguishing between the storage of facts and whether they are extractable in practice. With controlled experiments on synthetic biographies, the authors showed that training on diverse or augmented forms of facts (such as paraphrasing or sentence re-ordering) is crucial for facts to be recoverable via prompting. This is done with the use of linear probing to diagnose where facts are encoded in hidden states. Part 3.2 – Knowledge Manipulation~\citep{physicsllmsPart3.2} extends this discussion beyond storage and extraction by evaluating whether LLMs can flexibly use stored facts for downstream reasoning. Four factual manipulation tasks are defined (retrieval, classification, comparison, and inverse search) to show that LLMs, including GPT-4, are competent only at direct retrieval. Lastly, Part 3.3 – Knowledge Capacity Scaling Laws~\citep{allen_zhu2025physics} quantifies the knowledge storage limits of LLMs, demonstrating a consistent capacity of ~2 bits per parameter across various models. Their experimental design spans multiple architectures, quantization, and positional encoding methods.\\

\noindent \textit{Links between memorization capacity and model size.}
\citet{HowMuchKnowledgeCanYouPack} investigate the extent to which factual knowledge can be stored in model parameters by fine-tuning language models of varying sizes on closed-book QA tasks. They find that performance scales with the model size, with larger models being able to approach the accuracy of retrieval-based systems. \citet{allen_zhu2025physics} present exhaustive experiments in a controlled setting by fine-tuning models on synthetic biographical datasets under different conditions (training duration, model architectures, quantization, and data augmentations) and conclude that most models achieve a knowledge capacity of 2 bits per parameter regardless of quantization.\\

\noindent \textit{Wider works on factual memorization.}
Early empirical studies establish that pretrained LMs encode relational and factual knowledge. \citet{LLMS_as_KnowledgeBases} shows that BERT can answer cloze-style queries competitively with supervised baselines, even without fine-tuning. \citet{Radford2019LanguageMA} further demonstrates that autoregressive models like GPT-2 acquire zero-shot QA abilities through scale alone, with performance improving log-linearly with model size. Subsequent work formalizes these behaviors. \citet{wang-etal-2024-knowledge-mechanisms} propose a taxonomy of memorization, comprehension, and application, highlighting the fragility of in-weight knowledge. \citet{kadavath2022languagemodelsmostlyknow} show that, to some extent, LLMs can self-assess their factual correctness via self-evaluation metrics \( P(\text{True}) \) and \( P(\text{I Know}) \). Such estimations are found to improve with model size or with ensembles. These findings inform recent work on confidence modeling and selective generation. \citet{LearningForgettingRemembering}, the authors track training dynamics and find that factual memorization is highest early and late in training, with mid-training phases marked by forgetting. These observations have implications for curriculum learning and data curation strategies. Finally, \citet{LLMs_struggle_long_tail} links factual recall and the frequency of that information in pretraining corpora. The authors show that models underperform on rare facts, even at a large scale, pointing to a data coverage bottleneck and motivating retrieval-augmented approaches.\\

\noindent \textit{Knowledge extraction from LLMs.}
Prompt design plays a central role in retrieving stored knowledge. \citet{HowCanWeKnowWhatLLMSKnow} shows that standard close prompts underestimate LLM knowledge and proposes automated prompt mining and paraphrasing to improve extraction accuracy on the LAMA benchmark. Similarly, \citet{zhu2025physics_scaling_law} demonstrates that without sufficient factual augmentation, stored facts may remain unrecoverable, highlighting a gap between memorization and extractability. Both works stress that retrieval performance depends on prompt diversity and internal representations.\\

\noindent \textit{Learning novel knowledge and avoiding forgetting.}
LLMs face challenges when integrating new knowledge post-pretraining. The authors of~\citet{gekhman2024fine} demonstrate that fine-tuning on novel facts is both inefficient and destabilizing: new knowledge is learned slowly. It increases hallucination rates, suggesting that fine-tuning may conflict with pre-trained knowledge. The study in~\citet{PhysicsLLMsPart3.1} echoes this by showing that only sufficiently augmented knowledge is robustly encoded and extractable. The broader review in~\citet{HowDoLLMSCaptureChangingKnowledge} surveys continual learning methods, emphasizing the need for non-destructive updates that preserve anterior capabilities. Collectively, these works highlight the fragility of in-weight representations when adapting to novel information.\\

\noindent \textit{Knowledge editing.}
Another line of research considers the editing of stored knowledge in LLMs via direct mechanistic intervention on the weights, aiming to alleviate the computational burden of retraining. A wide range of editing techniques is reviewed by \citet{ComprehensiveStudyKnowledgeEditing}. \citet{KnowledgeEditingforLLMsSurvey} frame editing techniques in terms of cognitive-inspired mechanisms (external injection, model merging, and intrinsic editing), and introduce a unified benchmark (KnowEdit) alongside analyses of knowledge localization. \citet{huang2025can} tackle the core question of whether knowledge editing can correct hallucinations, introducing HalluEditBench and evaluating editing methods along multiple axes such as locality and generalization. Their results underscore both the promise and limitations of editing techniques in practice. Complementarily, Part 3.2 of the Physics of LLMs series \citep{physicsllmsPart3.2} highlights a deeper structural limitation: even when facts are faithfully stored in the model, their downstream manipulation (e.g., comparison or inverse search) remains brittle—suggesting that editing alone may not grant models flexible reasoning over internalized knowledge.\\

\noindent \textit{Mechanistic understanding of memorization.}
A series of recent works uncovers the mechanistic basis of how transformers store and recall information, with a particular focus on two mechanisms: induction heads, which enable in-context learning by attending to repeated structures within a prompt, and associative memories, where weight matrices implicitly store token associations via outer products. \citet{bietti2023birthtransformermemoryviewpoint} analyzes a simplified training setup, showing that transformers first learn global bigram statistics before gradually developing induction heads that enable pattern completion from local context. They also interpret certain weight matrices as associative memories formed during training. Building on this, \citet{LearningAssociativeMemories} provides a theoretical analysis of gradient-based learning in associative memory modules, revealing oscillatory learning dynamics caused by imbalanced token frequencies and correlated embeddings. These oscillations affect how quickly and reliably memories are stored. \citet{ScalingforAssociativeMemories} derive scaling laws for memory capacity and retrieval reliability, validating them empirically in small transformer models. Collectively, these works offer a theoretical and empirical foundation for how memorization emerges through optimization dynamics in LLMs.\\

\paragraph{\bf Tool-use in large language models.}
Retrieval-augmented generation (RAG)~\citep{RAGforKnowledgeIntensive} represents an early and influential line of work that treats external tools (e.g., retrievers) as memory extensions, enabling LLMs to access factual information without storing it parametrically. More broadly, the recent survey~\citep{ToolLearningSurvey} systematically categorizes tool-use into planning, selection, calling, and response stages, and highlights its promise for improving factuality, compositional reasoning, and generalization. A growing number of works explore how models acquire tool-use capabilities. Toolformer~\citep{schick2023toolformerlanguagemodelsteach} introduces a self-supervised setup where LLMs learn when and how to call APIs (e.g., calculators, search) by generating and filtering synthetic training data. MetaTool~\citep{MetaTool} introduces a benchmark to assess tool-use awareness and selection under realistic agent settings, showing that current LLMs still struggle with tool choice and reasoning under ambiguity. 
\citet{houliston2024uncertaintypenalizeddirectpreferenceoptimization} offers a principled offline preference-based RL algorithm tailored to LLM alignment under ambiguous context by leveraging reward or preference uncertainty estimates - such a method could naturally extend to tool learning, where ambiguity in delegation decisions is a central challenge. Combining tool use with multi-step reasoning, ART~\citep{ART-AutomaticReasoningTool} proposes a framework where LLMs write reasoning steps as executable programs that can call tools and integrate their outputs seamlessly. This work illustrates how tool use can serve as a functional memory system to complement and augment the limited and potentially brittle parametric memory of LLMs.

\newpage 
\section{Theoretical Details}
\label{app:theory}

In this section, we prove the bound from Theorem~\ref{thm:upper-bound} on the number of parameters needed for a transformer to learn when and how to query a retrieval system.
Using the same example and notations as in Section~\ref{sec:setting}, given a question of the form
\[Q = \underbrace{\text{Where was}}_{\phi_1(a)}\ \underbrace{\text{Thierry de Sienne}}_{\phi_2(n) = n}\ \underbrace{\text{born?}}_{\phi_3(a)},\] the model $f$ must output a well-formatted string
\begin{align*}
 T & = \underbrace{\text{\small To answer this request, I will make a tool-call.
\texttt{<DB>} FIND birthplace}}_{\chi_1(a)}\ \underbrace{\text{\small FOR Thierry de Sienne \texttt{</DB>}}}_{\chi_2(n)}, 
\end{align*}
where \verb|<DB>| and \verb|</DB>| are special tokens which, when parsed, result in the query being provided to the retrieval system.
Let us call this subtask the querying task.
The retrieval system then returns a string
\[
  \xi(f, a, n) = \text{``Germany''},
\]
containing the answer to $Q$. 
Then, given the context $Q,T,\xi(f, a, n)$, $f$ must return a well-formatted answer
\[
A = \underbrace{\text{Thierry de Sienne}}_{\psi_1(n)}\ \underbrace{\text{was born in}}_{\psi_2(a)}\ \underbrace{\text{Germany}}_{\psi_3(v)}.
\]
Let us call this second subtask the formatting task.

We only prove that the querying task, i.e., outputting $T$ as a function of $Q$, can be learnt by a transformer for which the bounds from Theorem~\ref{thm:upper-bound} apply: similar arguments are enough to prove that the formatting task can be learnt as well (and simultaneously) by such a model.

Furthermore, we make the following simplifying assumptions:
\begin{enumerate}
    \item[1.] The map 
    $\phi_1 : a \mapsto \phi_1(a)$ is injective.
    \item[2.] The only question mark in each question $Q$ is at the very end of the string.
    \item[3.] The string $Q$ is provided to the model's tokenizer in such a way that each substring $\phi_1(a), \phi_2(n), \phi_3(a)$ is separately tokenized, i.e., the tokenized version of $Q$ is such that no token contains characters belonging to several substrings.
    \item[4.] There does not exist $a, a' \in {\cal A}$ such that $a\neq a'$ but $\phi_1(a)$ is a substring of $\phi_1(a')$.
    \item[5.] The number of tokens in the questions and answers does not exceed a given constant.
    
\end{enumerate}
The first, second, and third assumptions are reasonable and could be relaxed at the cost of a more complex proof. 
Some variant of the fourth assumption cannot be avoided without additional constraints on the names $\cal N$, as two distinct pairs $(a,n), (a',n')$ could otherwise be mapped to the same question $Q$.
The last condition forces the number of tokens in the names, attribute,s and values to be less than some large number, e.g., 10k. This weak constraint is essentially a technicality linked to the limitations of positional embeddings, hence to the context length of any transformer, rather than a true algorithmic constraint.

\begin{thm}[Parameter Upper Bound, querying task]\label{thm:upper-bound_querying_task}
Consider a finite set $\cal A$ of attributes and a set $\cal N$ of names.
Then there exists a transformer with at most $8$ transformer blocks, an embedding dimension of at most $O(|\cal A|)$, and a total number of parameters at most $O(|{\cal A}|^2)$, which can achieve perfect accuracy on the querying task, i.e., that outputs
\[T = \chi_1(a) \circ \chi_2(n)\]
as a function of 
\[Q = \phi_1(a) \circ \phi_2(n) \circ \phi_3(a)\]
for any $a\in {\cal A}, n\in {\cal N}$.
\end{thm}

\begin{proof}
We focus on the most informative elements of the proof and omit some tedious details.

\noindent 
\begin{minipage}[c]{0.65\textwidth}
Given a transformer with $L$ layers and $k\in\{1,\ldots,L\}$ and internal dimension $D$, we let $a_k \in \R^D $ denote the input of the feedforward network of the $k$-th transformer block (i.e. the renormalization of the sum of the output of the multi-head attention and the residual connection), and $y_k\in \R^D$ denote the output of the $k$-th transformer block (i.e. the renormalization of the sum of the output of the feedforward network and of the block's second residual connection, see Figure~\ref{fig:transformer} for an illustration of a Transformer block).
\end{minipage}%
\hfill
\begin{minipage}[c]{0.3\textwidth}
    \centering
    \includegraphics[width=0.4\textwidth]{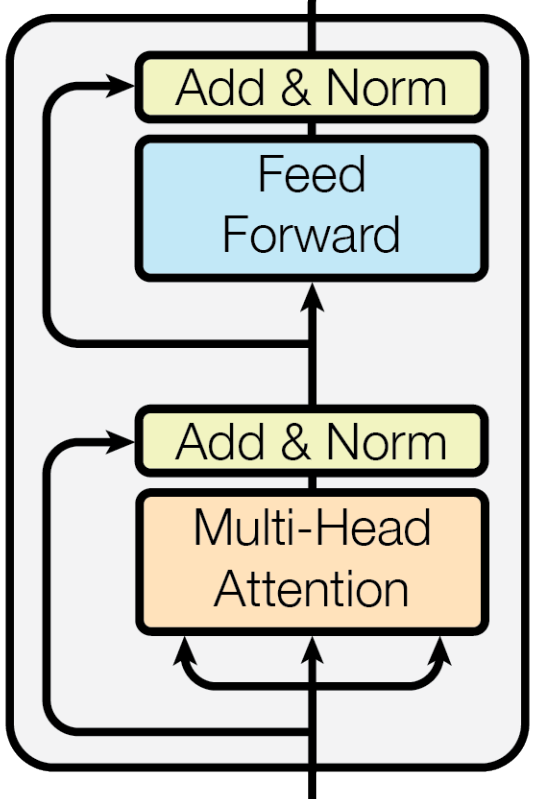}
    \captionof{figure}{Architecture of a Transformer block, \citet{vaswani2017attention}.}
    \label{fig:transformer}
\end{minipage}

\medbreak 

In what follows, we mean by ``the vector $a_k$ above the $i$-th token of sequence $S$'' the vector $a_k$ found within the $k$-th layer of the transformer and to which a direct line of residual connections starting at the embedding of the $i$-th token leads (and same for $y_k$).

By ``an encoding of $X$'', for some quantity or object $X$ that can take values in some set $S$, we mean the image of $X$ by some injection $I: S \rightarrow \R^n$.

Committing a minor abuse of conventions, we say that a vector $v\in \R^m$ (e.g. $a_k$) ``contains'' another vector $u\in \R^n$  (e.g. an encoding of some object) if the coordinates $u$ are a subset of the coordinates of $v$, i.e. $u = \pi(v)$ for some orthogonal projection along a sub-family of the orthogonal basis.
By convention, whenever we state that a vector $v$ contains several vectors $u_1,\ldots, u_l$, the dimensions in which those vectors are contained do not intersect, i.e. $v = \{u_1\}\times \ldots \times \{u_l\} \times \{w\}$ for some vector $w$ and up to some permutation of its coordinates.

When we say that a vector $a_k$ or $y_k$ above a given token contains ``a flag'' corresponding to a certain condition on the token and its predecessors, we mean that one of the coordinates of the vectors (without overlap with the coordinates used to store other information) is equal to $1$ if this condition is fulfilled (the flag is raised), and to $0$ otherwise.

Remember that we must design a transformer that outputs the sequence 
\begin{align*}
 T & = \underbrace{\text{\small To answer this request, I will make a tool-call.
\texttt{<DB>} FIND birthplace}}_{\chi_1(a)}\ \underbrace{\text{\small FOR Thierry de Sienne \texttt{</DB>}}}_{\chi_2(n)}, 
\end{align*}
when provided with
\[Q = \underbrace{\text{Where was}}_{\phi_1(a)}\ \underbrace{\text{Thierry de Sienne}}_{\phi_2(n) = n}\ \underbrace{\text{born?}}_{\phi_3(a)}.\]
Concretely and due to the auto-regressive nature of transformers, the model will progressively concatenate tokens to $Q$ until a joint string $Q \circ T$ has been produced.

We are going to show that there exists a transformer with $8$ layers and embedding dimension $C|\cal A|$ for $C>0$ large enough such that for any input question $Q = \phi_1(a) \circ \phi_2(n) \circ \phi_3(a)$ (for some $a\in {\cal A}, n\in {\cal N}$), the following facts hold simultaneously\footnote{In the proof, approximating functions, for instance needed to compute the absolute position, amounts to a perfect approximation on a finite subset using the fact that the set of tokens is discrete}.


$\blacktriangleright$ \textbf{Fact: any $y_k$ above any token contains an encoding of the absolute position of the token within the entire sequence.}
The positional encodings, which contain the desired information, are added to the tokens' learned embedding to form the input to the model's first transformer block. For a large enough embedding dimension and a judicious choice of learned embeddings, and as long as the number of tokens does not exceed a given limit, the absolute position can be recovered by the first transformer block (using the fact that two-layer neural networks are universal approximators and assuming again that the embedding dimension is large enough).
This information can then be stored in some of the coordinates of $y_1$, and be passed to all consecutive $y_k$ (e.g. using the residual connections and trivial attention heads).
 Note that the embedding dimension needed does not depend on $\cal A$ or $\cal N$.

$\blacktriangleright$ \textbf{Fact:  $y_3$ contains one flag per attribute $a'\in {\cal A}$. The flags corresponding to $a' \neq a$ are never raised, and the flag corresponding to $a$ is only raised above the last token of $\phi_1(a)$.}

At any step, let $t$ denote the last token of the currently processed sequence, and let $p$ be its absolute position.
As stated above, $y_1$ contains an encoding of $p$. It can also contain an encoding of $t$ (using similar arguments).

The attention layer of the second transformer can then pass this information to its feedforward network. Using again the approximation properties of neural networks, for each $a'\in {\cal A}$ there can be a dedicated entry $v_{a'} \in \R$ in the network's output $y_2$ such that:
\begin{itemize}[noitemsep]
    \item if $p>\text{len}(\phi_1(a))$, then $v_{a'} = -2$,
    \item if $p\leq \text{len}(\phi_1(a))$ and $t$ is not equal to the $p$-th token of $\phi_1(a)$, then $v_{a'} = -2$.
    \item if $p\leq \text{len}(\phi_1(a))$ and $t$ is equal to the $p$-th token of $\phi_1(a)$, then $v_{a'} = 1/\text{len}(\phi_1(a))$ and 
\end{itemize}

In the third attention layer, the coordinates $v_{a'}$ corresponding to each $a' \in {\cal A}$ can be summed. If $a \neq a'$, this sum is strictly negative. If $a = a'$, then:
\begin{itemize}[noitemsep]
    \item if $p<\text{len}(\phi_1(a))$, then the sum is positive but strictly smaller than $1$,
    \item if $p = \text{len}(\phi_1(a))$, the sum is exactly $1$, and
    \item if $p>\text{len}(\phi_1(a))$, the sum is strictly negative.
\end{itemize}
Hence, all cases are differentiated. The following feedforward network can then output within $y_3$ a flag for all $a'\in {\cal A}$, which is raised only if    $a = a'$ and the current token is the last token of $\phi_1(a)$.
As the lengths of the substrings $\{\phi_1(a')\}$ are bounded, this can be done with an embedding dimension in $O(|\cal A|)$.\\

$\blacktriangleright$ \textbf{Fact: $y_1$ contains a flag that is only raised above the question mark at the end of $\phi_3(a)$, and another flag that is raised above every token after the question mark.  }

This is essentially trivial (using our second assumption).\\

$\blacktriangleright$ \textbf{Fact: $y_4$ contains a vector that encodes the length in tokens of $n$ above every token after the question mark at the end of $\phi_3(a)$.}

The facts above state that $y_3$ contains an encoding of the absolute position of the current token, a flag $f_a$ corresponding to $a$ that is only raised above the last token of $\phi_1(a)$, a flag $f_?$ that is only raised above the question mark at the end of $\phi_3(a)$, and a flag $f_{\geq ?}$ that is raised above every token after the question mark (these two flags are passed from $y_1$ to $y_3$).

At each token, two non-trivial queries are made in the fourth attention layer if and only if $f_{\geq ?}$ is raised. The first non-trivial query matches only with a key generated from $f_a$ and its output value is an encoding of the position of the last token of $\phi_1(a)$ as well as an encoding of the value of $a \in \cal A$. The second non-trivial query matches only with a key generated from $f_?$ and its output value is an encoding of the position of the question mark at the end of $\phi_3(a)$.
The fourth feedforward network then takes these two positional encodings and the identity of $a\in \cal A$, and outputs (in some subset of the coordinates of $y_4$) the distance between the last token of $\phi_1(a)$ and the question mark, minus the length in tokens of $\phi_3(a)$. This is precisely the length in tokens of $n$.
This can be done with an embedding dimension in $O(|\cal A|)$.\\

$\blacktriangleright$ \textbf{Fact: the model successfully outputs $\chi_1(a) \circ \text{``FOR''} $ after $Q$. }

A non-trivial query is made in the second attention layer if and only if the flag $f_{\geq ?}$ is raised. This query matches the flag $f_?$ and brings to $y_2$ an encoding of the position of the question mark. $y_2$ also contains an encoding of the position of the current token, and a copy of the flag $f_{\geq ?}$.
This information is passed to $y_3$.
The fourth attention layer can match the flag above the last token of $\phi_1(a)$ which describes the value of $a\in \cal A$ and pass this value to the fourth feedforward network, along with the encoding of the position of the question mark, the encoding of the position of the current token and the copy of the flag $f_{\geq ?}$.
If $f_{\geq ?}$ is raised, then the fourth feedforward network can compute the distance $l$ between the question mark and the current token, and output an encoding of the $l$-th token of $\chi_1(a) \circ \text{``FOR''} $ (or a flag saying that no token should be output if $l$ is larger than the length of $\chi_1(a) \circ \text{``FOR''} $).
This can be done with an embedding dimension in $O(|\cal A|)$.
This encoding can then be passed to the following layers, until the correct $l$-th token is output.\\

$\blacktriangleright$ \textbf{Fact: $y_5$ contains a flag that is only raised above the last token of  $\chi_1(a) \circ \text{``FOR''} $.}

This follows from the same arguments as the previous fact.\\

$\blacktriangleright$ \textbf{Fact: above every token after $\chi_1(a) \circ \text{``FOR''} $, $y_6$ contains an encoding of the distance between the current token and the last token of $\chi_1(a) \circ \text{``FOR''} $.}

The sixth attention layer can match with the flag from the previous fact to pass its position to the sixth feedforward layer, as well as the position of the current token. These can be processed by the feedforward layer to output the desired distance.\\

$\blacktriangleright$  \textbf{Fact: above every token after the last token of $\chi_1(a) \circ \text{``FOR''} $, $y_5$ contains an encoding of the distance $d$ between the first token of $\phi_2(n) = n $ and the last token of $\chi_1(a) \circ \text{``FOR''} $.}

This distance is equal to the length of $n$ plus the length of $\phi_3(a)\circ \chi_1(a) \circ  \text{``FOR''}$.
The length of $n$ is encoded in $y_4$ above every token after the question mark at the end of $\phi_3(a)$ (using an earlier fact). This quantity can then be passed to the fifth feedforward network.
$y_3$ contains a flag corresponding to $a$ above the last token of $\phi_1(a)$  (using an earlier fact again), which is raised nowhere else. This quantity can be passed to $y_4$, queried by a dedicated attention head in the fifth attention layer, and passed to the fifth feedforward network.
As there are finitely many $a'\in \cal A$, the fifth feedforward network can memorize every length $\phi_3(a')\circ \chi_1(a') \circ  \text{``FOR''}$ with an embedding dimension in $O(|\cal A|)$.
It adds the length of  $\phi_3(a')\circ \chi_1(a') \circ  \text{``FOR''}$ to that of $n$, and encodes it into $y_5$.\\

$\blacktriangleright$  \textbf{Fact: the model successfully outputs $n$ after  $\chi_1(a) \circ \text{``FOR''} $, and $y_7$ contains a flag that is only raised above the last token of (this copy of) $n$. }

For every token after $\chi_1(a) \circ \text{``FOR''} $, the sixth layer can use the encoding of the distance $d$ from the previous fact and the encoding of the position of the current token to output an encoding of the position $p$ of the token that is $d$ entries before the current token.
The seventh attention layer can then use this encoding to match the token at position $p$ (which is part of $n$) and get its value $t$. We know that $y_6$ also contains an encoding of the distance $d'$ between the current token and the last token of $\chi_1(a) \circ \text{``FOR''} $.
If $d'$ is smaller than the length of $n$, the seventh feedforward network outputs a copy of $t$, which lets the model output $t$.
If $d'$ is exactly the length of $n$ (i.e., the current token is the last token of the copy of $n$), the seventh feedforward network outputs a flag within $y_7$.
If $d'$ is greater than $n$, the flag is not raised.
Hence the model does output $n$ after  $\chi_1(a) \circ \text{``FOR''} $.\\

$\blacktriangleright$ \textbf{Fact: the model successfully outputs \texttt{</DB>} after $n$, and thus successfully outputs $T$ in response to $Q$.}

The eighth attention layer matches the flag in $y_7$ that is raised above the last token of $n$. It passes the existence of this flag, the position of the last token of $n$, and the position of the current token to the eighth feedforward network, which computes the distance $d''$ between the two positions and outputs an encoding of the $d''$-th token of \texttt{</DB>}, which the model can then output.

The combination of these facts is enough to complete the proof of Theorem~\ref{thm:upper-bound_querying_task}.
\end{proof}

\newpage 
\section{Experimental Details}
\label{app:experiments}
In this section, we provide details on our experimental setup, specifically regarding the database construction, language modeling pipeline, and training runs. We also provide additional results in both controlled and large-scale settings. Seeds, when provided, are relative to distinct data generation and network initialization, and as such, results are presented with the standard deviation across runs (e.g., we conducted 10 different runs to obtain Figure~\ref{fig:controlled_scaling}).

\paragraph{\bf Reproducibility.} The code and reproducibility instructions can be found at \code{}. We provide in the main paper and the appendix all the details needed to reproduce our results. Experiments were conducted using high-performance GPUs such as NVIDIA V100 and A100. Our implementation supports distributed training, but we have also ensured that it can be run on a single device. 

\subsection{Factual Recall Database}
\label{app:data}
In our experiments, we construct synthetic biographical datasets using a fixed list of names ${\cal N}$ (each name is a pair \{first name-last name\}) and four attributes ${\cal A}$: ``birth place'', ``birth date'', ``current address'', ``occupation''; which can take 7, 16800, 213, and 100 values respectively. This yields $4|{\cal N}|$ atomic facts per dataset instance. We present in Figure~\ref{lst:template} an example of such a dataset, where we query the birth date of Kenny McRoy, comparing the \emph{in-weight} and \emph{in-tool} settings. We adopt a dialog format, with a chat template. We design our tokenizer to take care of the special tokens needed (see Appendix~\ref{app:controlled_exp}). We can see that the in-weight setting leads to a single-turn dialog for the LLM while the in-tool one requires a multi-turn dialog.
\begin{figure}[!h]
\small
\begin{center}
\begin{minipage}{\linewidth}
\begin{lstlisting}[language=Python, frame=single]
# In-weight query
{
    "dialog": [
        {"source": "user", "content": "When was Kenny McRoy born?"},
        {"source": "assistant", "content": "Kenny McRoy was born on 19/05/1998."},
    ],
    "people_id": 0,
    "answer": "19/05/1998",
}

# In-tool query
{
    "dialog": [
        {"source": "user", "content": "When was Kenny McRoy born?"},
        {
            "source": "assistant",
            "content": "To answer this question I will make a request to the database:\n```sql\nFIND birth_date FOR Kenny McRoy\n```",
        },
        {"source": "database", "content": "19/05/1998"},
        {"source": "assistant", "content": "Kenny McRoy was born on 19/05/1998."},
    ],
    "people_id": 0,
    "answer": "19/05/1998",
}
\end{lstlisting}
\end{minipage}
\end{center}
\caption{Examples of queries for the in-weight versus and in-tool settings.}
\label{lst:template}
\end{figure}

\newpage 
\subsection{Controlled Experiments}
\label{app:controlled_exp}
In our experiments of Section~\ref{sec:controlled_exp}, we train small Llama3-style language models~\citep{grattafiori2024llama3herdmodels} from scratch. To that end, we build a language modeling pipeline following the literature standard~\citep{meta_lingua}. Below, we provide details on the tokenization, the architecture, and the training setup, alongside further results.

\paragraph{\bf Tokenizer.} Transformer models~\citep{vaswani2017attention}, like most neural networks, operate on numerical data, and in particular, sequences of vectors. To perform language modeling, a given sequence is mapped to a sequence of numerical vectors by first splitting it into words or subwords and then encoding each word or subword into a vector. This process is referred to as the tokenization process and is typically performed using tokenizers. Most state-of-the-art language models have an associated pre-trained tokenizer; it is also possible to encode text using their byte representation. Since we pretrain our model from scratch, we considered the latter byte tokenizer to avoid potential bias from pretrained tokenizers. It simply splits words into single-character tokens and maps each of them to their raw byte representation (taking values in $[0,256)$). It results in a vocabulary space of size $256$ for the byte encoding ($256 = 2^8)$, which can be augmented by adding special tokens. The vocabulary size reaches $260$ with the addition of $4$ special tokens: \textbf{\color{\usercol} \textit{<user>}}, \textbf{\color{\llmcol} \textit{<assistant>}}, \textbf{\color{\datacol} \textit{<database>}} and \textbf{\color{\eodcol} \textit{<eod>}}. Each plays a specific role during the dialog between a user and a language model assistant:
\begin{itemize} 
\setlength\itemsep{1.5pt}
    \item The \textbf{\color{\usercol} \textit{<user>}} token marks the beginning of the dialogue by the user, 
    \item The \textbf{\color{\llmcol} \textit{<assistant>}} token marks the language model assistant turn, which can either answer directly or make a call the the tool,
    \item The \textbf{\color{\datacol} \textit{<database>}} token marks the beginning of the database result to the request,
    \item The \textbf{\color{\eodcol} \textit{<eod>}} token indicates the end of the dialog.
\end{itemize}

\paragraph{\bf Chat templates.} We model the factual recall task as a dialog between a user and a language model assistant. The user asks questions about a given person, for instance, their date of birth, birthplace, occupation, or current address. The assistant can then either answer directly using the information contained in its weights (in-weight learning) or make a call to a tool, implemented in our code as an SQL agent that has access to a database of biographies of persons. To decode such interactions, we implement a wrapper around the tokenizer to allow for dialog interactions, akin to the HuggingFace chat templates.

\paragraph{\bf Online generation of batches.} Data is organized into subsets that each represent a source of data. It follows the standard in the literature, which accounts for the fact that, in practice, pretraining data aggregates texts from different sources like Wikipedia or arXiv. To optimize data generation, batches are created on the fly by sampling with pre-defined proportions from the different subdirectories. Text is tokenized also on the fly by iterating through each line of the sampled JSONL files.

\paragraph{\bf Transformer architecture.} Our models are small transformer decoders following the Llama3 implementation~\citep{grattafiori2024llama3herdmodels} with Rotational Positional Embedding~\citep{su2024rope}, Flash Attention~\citep{dao2022flashattention}, efficient attention for prefilling and token generation, gated linear units~\citep{dauphin2017glu, shazeer2020gluvariantsimprovetransformer} with a SiLU activation~\citep{hendrycks2023gaussianerrorlinearunits}, RMSNorm layers~\citep{zhang2019rms}, and KV caching~\citep{pope2022efficientlyscalingtransformerinference} for the inference. The feed-forward hidden dimension is fixed to $4$ times the embedding dimension. For each dataset size, we train a family of small Llama3-style transformer models~\citep{grattafiori2024llama3herdmodels} with 2 layers, 2 attention heads, a vocabulary size of 300, and a context window of 257. 
The embedding dimension ranges from 4 to 128 (to ensure that ROPE constraints between the number of heads and the embedding dimension are met), resulting in models with between 2K and 0.6M parameters. 

\paragraph{\bf Training.} Our models are trained for up to $100{,}000$ steps using the AdamW optimizer~\citep{loshchilov2018decoupled}, with a batch size of $128$ samples, a decoupled weight decay coefficient of $0.1$, and $(\beta_1, \beta_2) = (0.9, 0.95)$.
We use a cosine learning rate scheduler with a warmup phase of $50$ steps and a maximum learning rate of $0.001$, and a final learning rate ratio of $0$.

\paragraph{\bf Evaluation and inference.} At inference time, we need to generate tokens sequentially, which is known to be slow. To optimize it, we make use of the standard practices in the literature, like KV caching. To evaluate the factual recall of our model, we feed it prompts to be completed. We implement additional attention modes, such as prefilling and generation, to decode the dialog and allow interacting with the agent. These operations are parallelized by defining a specific prompt loader during evaluation. It efficiently deals with queues of prefilled tokens to generate prompt completions. 

\paragraph{\bf Tool implementation.} In our work, the tool is an SQL agent relying on \texttt{sqlite3} to access a database containing biographies of people with their first name, last name, date of birth, place of birth, occupation, and current address. The language model assistant can learn to make queries to this database thanks to in-tool learning. The SQL agent can be used to create the database and insert new elements in it, and has agentic capabilities not only to execute a query and return a result but also to parse instructions from an LLM prompt, execute them, and answer in natural language to the language model assistant.

\begin{figure}[!h]
    \centering
    \includegraphics[width=0.5\linewidth]{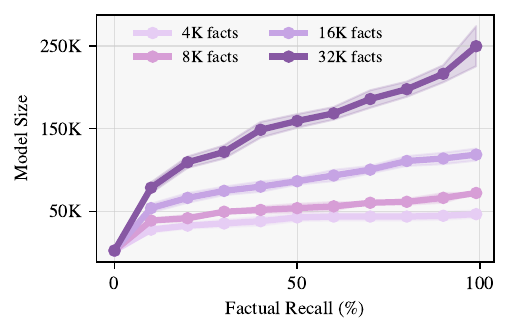}
    \caption{\textbf{Scaling of parameter requirements with the level of factual recall.} 
    Minimum number of parameters required as a function of the factual recall. The higher the number of facts retrieved, the bigger the models. When the total number of facts to retrieve increases, we observe that the parameter requirement increases too. This shows that in-tool learning is all the more beneficial when very effective models are needed on large databases.}
    \label{fig:recall}
\end{figure}

\paragraph{\bf Further results.} In the experiments displayed in Section~\ref{sec:controlled_exp}, the recall accuracy is fixed at $95\%$. To better understand how it can impact the parameter requirement, we display in Figure~\ref{fig:recall} the minimum parameter requirement as a function of the factual recall for 4 different numbers of facts. We notice that the higher the recall, the bigger the models, and that this increase is steeper when the number of facts to retrieve increases. This can be explained from Theorem~\ref{thm:lower-bound}, which makes the dependency between the model size and the number of facts explicit. Given that tool use enables the retrieval of an unbounded number of facts without the need to increase the number of parameters, as shown in Figure~\ref{fig:controlled_scaling}, this implies that in-tool learning is all the more beneficial when one wants a more effective model on a bigger dataset. This follows the increasing trends of toll use in the literature for harder benchmarks such as mathematics.

\newpage 
\subsection{Large Scale Experiments}
\label{app:large_scale_exp}
In this section, we provide additional results related to Section~\ref{sec:large_scale_exp}.

\begin{figure}[H]
\centering
\begin{minipage}[t]{.485\textwidth}
  \centering
  \includegraphics[width=\linewidth]{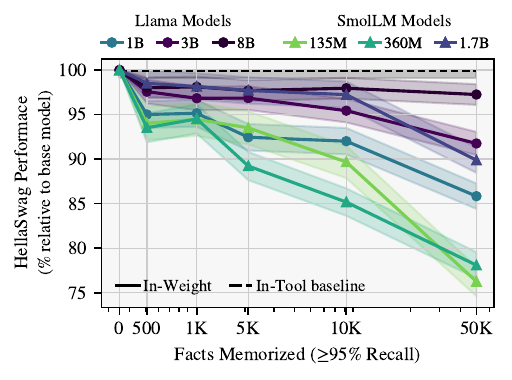}
  \caption{\textbf{HellaSwag performance (relative to base model) versus memorization load.} Same setup as Figure~\ref{fig:largescale_capability}; the dashed line represents the worst (lowest) performance among tool-models (Smol-135M).}
  \label{fig:appendix_largescale_hellaswagrelative}
\end{minipage}%
\hfill
\begin{minipage}[t]{.48\textwidth}
  \centering
  \includegraphics[width=\linewidth]{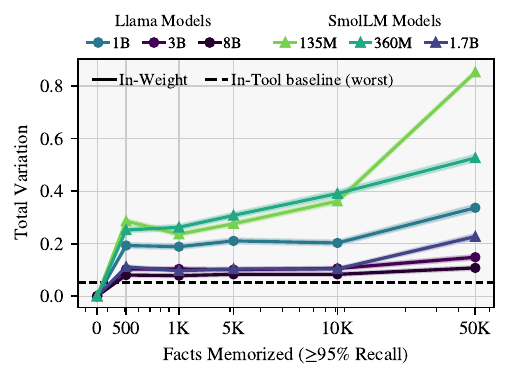}
  \caption{\textbf{Total Variation versus memorization load.} Same setup as Figure~\ref{fig:largescale_distance}; the dashed line represents the worst (highest) TV distance attained among tool-models (Smol-135M).}
\label{fig:appendix_largescale_TV_versus_MemoLoad}
\end{minipage}
\label{fig:benefits_samformer}
\end{figure}

\paragraph{\bf HellaSwag performance relative to base model.} 
Figure~\ref{fig:largescale_capability} in the main section of the paper displays the absolute HellaSwag accuracy attained by different models finetuned on datasets of increasing size. To complement this, we examine the relative drop in HellaSwag performance from the base model as the memorization load increases, in Figure~\ref{fig:appendix_largescale_hellaswagrelative}. This perspective offers a clearer view of the dynamics by which larger models maintain their prior capabilities under heavier memorization loads. Tool-learning (dashed line representing the worst performance over all tool-models) preserves general capabilities almost perfectly across model and dataset sizes. In contrast, in-weight learning leads to noticeable degradation, especially for smaller models and larger factual loads. This supports our theoretical prediction that parameter-based memorization causes interference and loss of prior capabilities, due to finite capacity and update interference. Larger models are more robust to such forgetting, but still exhibit performance decay beyond 10K facts.\\

    
A simplified intuition helps interpret the consistent ordering across model sizes: larger models possess more "capacity volume," so absorbing a fixed amount of new information replaces a smaller fraction of prior knowledge. While this view captures the trend, it likely omits key confounding factors: larger models are typically trained with more compute and data, and more complex architectures may also be more brittle to local updates—suggesting the optimization dynamics themselves differ in important ways. \\

\paragraph{\bf Total variation versus memorization load.} The intuition laid out above is also observed in the TV distance suffered by models as the memorization load grows: larger models diverge less than smaller counterparts. 
Figure~\ref{fig:appendix_largescale_TV_versus_MemoLoad} depicts the Total Variation of finetuned models from their base-model, for different memorization loads. The figure shows the same results as Figure~\ref{fig:largescale_distance} for in-weight models but replaces in-tool data by the highest (worst) TV achieved (by SmolLM 135M), represented as the dashed baseline. We notice in-tool TV baseline is constant across dataset sizes, and is lower than in-weight learning deviations - this is due to tool-use mastery unlocking unbounded factual recall. Overall, bigger models deviate less from their base model for the same in-weight memorization load. The distance is dramatically greater for smaller models on bigger datasets close to 50k facts.

\begin{figure}[!h]
    \centering
    \includegraphics[width=\linewidth]{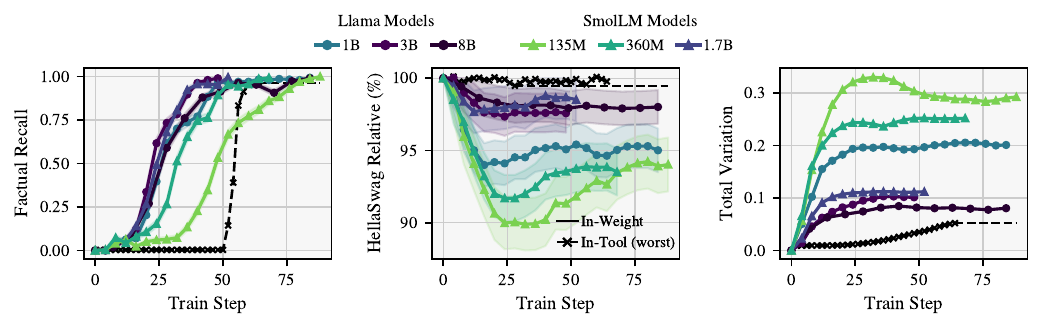}
    \caption{Metrics throughout training on a database of 500 facts. The dashed line represents the worst in-tool learning baseline, highlighting that tool use attains full recall while maintaining a very high $(\geq98\%)$ level of prior capabilities on HellaSwag, and by deviating less than $0.04$ in TV from its base model.}
    \label{fig:appendixlargescale_500}
\end{figure}

\begin{figure}[!h]
    \centering     \includegraphics[width=\linewidth]{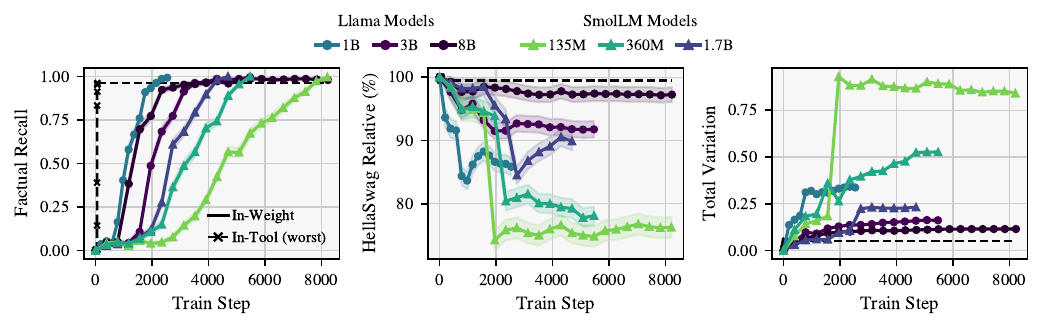}
    \caption{Metrics throughout training on a dataset of 50K facts. The worst in-tool model (dashed) mastered the tool in very few training steps, conserving higher HellaSwag capabilities and deviating less from its base model.}
\label{fig:appendixlargescale_50}
\end{figure}

\paragraph{\bf Training dynamics.} For databases of 500K and 50K facts, we display the recall accuracy, the HellaSwag performance, and the total variation attained by checkpoints throughout training until complete memorization in Figures~\ref{fig:appendixlargescale_500} and~\ref{fig:appendixlargescale_50}. Across model scales, we observe that recall accuracy improves rapidly during early training, with most models nearing saturation within 30--40 steps. However, a substantial number of additional steps are spent refining the final few percent of recall, suggesting that later training primarily sharpens the output distribution rather than acquiring new associations. Notably, most of the degradation in HellaSwag performance and increase in total variation occur early in training, indicating that the bulk of distributional shift and capability loss is incurred in the initial stage of training for memorization, while later steps have comparatively minor effects.


\end{document}